\documentclass{article}

\usepackage{arxiv}

\usepackage[utf8]{inputenc} 
\usepackage[T1]{fontenc}    
\usepackage{hyperref}       
\usepackage{url}            
\usepackage{booktabs}       
\usepackage{amsfonts}       
\usepackage{nicefrac}       
\usepackage{microtype}      
\usepackage{lipsum}		
\usepackage{graphicx}
\usepackage[numbers]{natbib}
\usepackage{doi}
\usepackage{amsmath,amssymb}
\usepackage{placeins}

\title{Vision Transformers for femur fracture classification}

\author{ \href{https://orcid.org/0000-0002-8813-6388}{\includegraphics[scale=0.06]{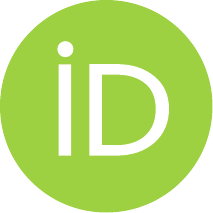}\hspace{1mm}Leonardo ~Tanzi}\thanks{\textit{Corresponding author.}} \\
    DIGEP\\
	Polytechnic University of Turin\\
	Turin, IT \\
	\texttt{leonardo.tanzi@polito.it} \\
	\And
	\href{https://orcid.org/0000-0002-1809-5058}{\includegraphics[scale=0.06]{orcid.pdf}\hspace{1mm}Andrea ~Audisio} \\
	School of Medicine\\
	University of Turin \\
	Turin, IT \\
	\texttt{andrea.audisio@unito.it} \\
	\And
	\href{https://orcid.org/0000-0002-2894-4164}{\includegraphics[scale=0.06]{orcid.pdf}\hspace{1mm}Giansalvo ~Cirrincione} \\
	LTI\\
	University of Picardie Jules Verne\\
	Amiens, FR \\
	\texttt{exin@u-picardie.fr} \\
		\And
	\href{https://orcid.org/0000-0002-2514-4719}{\includegraphics[scale=0.06]{orcid.pdf}\hspace{1mm}Alessandro ~Aprato} \\
	School of Medicine\\
	University of Turin \\
	Turin, IT \\
	\texttt{alessandro.aprato@unito.it} \\
		\And
	\href{https://orcid.org/0000-0001-8910-7020}{\includegraphics[scale=0.06]{orcid.pdf}\hspace{1mm}Enrico ~Vezzetti} \\
	DIGEP\\
	Polytechnic University of Turin\\
	Turin, IT \\
	\texttt{enrico.vezzetti@polito.it} \\
}

\renewcommand{\shorttitle}{Preprint}

\hypersetup{
pdftitle={Vision Transformers for femur fracture classification},
pdfsubject={ViT},
pdfauthor={L ~Tanzi, A ~Audisio, G ~Cirrincione, A ~Aprato, E ~Vezzetti},
pdfkeywords={Deep Learning, Vision Transformer, Femur Fracture, Self-Attention, CAD System},
}

\begin{document}
\maketitle

\begin{abstract}
\textit{Objectives}: In recent years, the scientific community has focused on the development of Computer-Aided Diagnosis (CAD) tools that could improve bone fractures’ classification, mostly based on Convolutional Neural Network (CNN). However, the discerning accuracy of fractures’ subtypes was far from optimal. This paper proposes a modified version of a very recent and powerful deep learning technique, the Vision Transformer (ViT), outperforming CNNs based approaches and consequently increase specialists’ diagnosis accuracy.  \\
\textit{Methods}: 4207 manually annotated images were used and distributed, by following the AO/OTA classification, in different fracture types, the largest labeled dataset of proximal femur fractures used in literature. The ViT architecture was used and compared with a classic CNN and a multistage architecture composed by successive CNNs in cascade. To demonstrate the reliability of this approach, 1) the attention maps were used to visualize the most relevant areas of the images, 2) the performance of a generic CNN and ViT was compared through unsupervised learning techniques and 3) 11 specialists were asked to evaluate and classify 150 proximal femur fractures’ images with and without the help of the ViT, then results were compared for potential improvement.  \\
\textit{Results}: The ViT was able to correctly predict 83\% of the test images. Precision, recall and F1-score were 0.77 (CI 0.64-0.90), 0.76 (CI 0.62-0.91) and 0.77 (CI 0.64–0.89), respectively. The average specialists’ diagnostic improvement was 29\% when supported by ViT’s predictions, outperforming the algorithm alone.  \\
\textit{Conclusions}: This paper showed the potential of Vision Transformers in bone fracture classification. For the first time, good results were obtained in sub-fractures classification, with the largest and richest dataset ever, outperforming the state of the art. Accordingly, the assisted diagnosis yielded the best results, proving once again the effectiveness of a coordinate work between neural networks and specialists.

\end{abstract}

\hspace{2pt} 

\newpage

\section{Introduction}

Musculoskeletal diseases represent the most common cause for severe, long-term disability worldwide \cite{woolf_burden_2003}. Due to the progressive aging of the population, the prevalence and incidence of fragility fractures is increasing and will continue so in the future \cite{reginster_osteoporosis_2006}. In 2010 the estimated incidence of hip fractures was 2.7 million patients per year globally. In the Emergency Department, a pelvis radiograph is performed when a hip fracture is suspected \cite{parker_hip_2006}. The correct evaluation and classification of proximal femur fractures by radiologists strongly affects future patients’ treatment and prognosis. The AO/OTA classification is hierarchical and provides a well-defined methodology for assessing fractures correctly, enabling physicians to guide treatment and communicate with a standardized language \cite{journal_of_orthopaedic_trauma_femur_2018}. However, the correct classification of hip fractures can be demanding for osteopenia, superimposition of soft tissues in obese patients and difficult patients’ positioning resulting in poorer image quality \cite{kirby_radiographic_2010}. The above mentioned difficulties, the stressful working environment of Emergency Departments and the perceived complexity of the classification might have affected its widespread utilization, limiting evidence-base fracture management and data collection for research.
In this context, implementing a CAD (Computer Assisted Diagnosis) system in doctors’ workflow might have a direct impact in patients’ outcome. This idea was demonstrated in a previous work \cite{tanzi_hierarchical_2020} from our research group, where a deep learning \cite{lecun_deep_2015} based method was used to classify femur fractures and the performance of physicians with and without its help was compared. Deep learning is becoming more and more widely used, giving astonishing results in different fields of application, such as surgery \cite{tanzi_intraoperative_2020, twinanda_endonet_2017, tanzi_real-time_2021} and face recognition \cite{olivetti_deep_2020}. In the vision domain, after the introduction of AlexNet \cite{krizhevsky_imagenet_2017} on the ImageNet competition in 2017, the applications of Convolutional Neural Networks (CNNs) have been increasing for their ability to capture the spatial dependencies in an image. Nevertheless, CNNs have different limitations. 
In recent times, a new paradigm called Transformer, introduced formerly for Natural Language Processing (NLP) \cite{vaswani_attention_2017}, has demonstrated exemplary performance on a broad range of language tasks, by means of BERT (Bidirectional Encoder Representations from Transformers) \cite{devlin_bert_2019} or GPT (Generative Pre-trained Transformer) \cite{radford_improving_2018}. Transformer architectures are based on a self-attention mechanism that learns the relationships between the elements of a sequence and 1) can deal with complete sequences, thus learning long-range relationships 2) can be easily parallelized 3) can be scaled to high-complexity models and large-scale datasets. The discovery of Transformer networks in the NLP domain has aroused great interest in the computer vision community. However, visual data follow a typical structure, thus requiring new network designs and training schemes. As a result, different authors have proposed their own implementation of a Transformer model applied to vision, obtaining great results in object detection \cite{carion_end--end_2020}, segmentation \cite{ye_cross-modal_2019}, video analysis \cite{girdhar_video_2019}, image generation \cite{zhang_self-attention_2019}, and many more.
In a previous publication from our group \cite{tanzi_x-ray_2020}, some selected papers concerning the topic of femur fracture classification have been reviewed, from basic approaches to main advanced solutions. Initial prior works for detection and classification of fractures \cite{cao_fracture_2015, myint_analysis_2018} focused on conventional machine learning processes consisting of pre-processing, feature extraction and classification phases. Recently, impressive results have been obtained using CNNs. Unfortunately, the majority of the existing works regarding fractures classification, focused mainly on the binary classification between \textit{Broken} and \textit{Unbroken bone}s \cite{lindsey_deep_2018, olczak_artificial_2017, rajpurkar_mura_2018}, which has a low impact on doctor diagnosis. To the best of our knowledge, pure CNNs have been applied to classify different types of fractures just in two previous papers \cite{chung_automated_2018, jimenez-sanchez_towards_2019}. More complex architectures were applied in \cite{lee_classification_2020}, where an encoder with a LSTM (Long Short Term Memory) layer obtained very good results even in sub fracture classification, but using also radiology reports which are not always easy to obtain, and in \cite{kazi_automatic_2017}, where the authors used spatial transformer and a CNN to localize and classify the fractures. Nevertheless, results are still non optimal, especially for complex fractures, and a generalized approach does not exist yet. \\
The novelties introduced by this work are four-fold:
\begin{enumerate}
    \item the largest and richest labeled dataset ever for femur fractures classification was used, with 4207 images divided in 7 different classes;
    \item the Vision Transformer (ViT) implementation by Google \cite{dosovitskiy_image_2021} was applied for the classification task, surpassing the two baselines of a classic InceptionV3 \cite{szegedy_rethinking_2016} network and a hierarchical network proposed in \cite{tanzi_hierarchical_2020}. This is the first work where CNN were not involved in the classification pipeline;
    \item 	the attention maps of ViT were visualized and the output of the Transformer’s encoder was clustered in order to understand the potentiality of this architecture;
    \item 	a final evaluation was carried out, asking to 11 specialists to classify 150 images by means of an online survey, with and without the help of our system. 
\end{enumerate}

This paper is structured as follows. Section 2 covers materials and methods that are relevant for the design of this approach.  Section 3 shows the overall performance and evaluation.  Section 4 discusses our findings, recommendations, future work and summarizes our conclusions.     

\section{Materials and methods}
\label{sec:headings}

\subsection{AO/OTA Proximal Femur Classification}
\begin{figure}
	\centering
	\includegraphics[width=.8\textwidth]{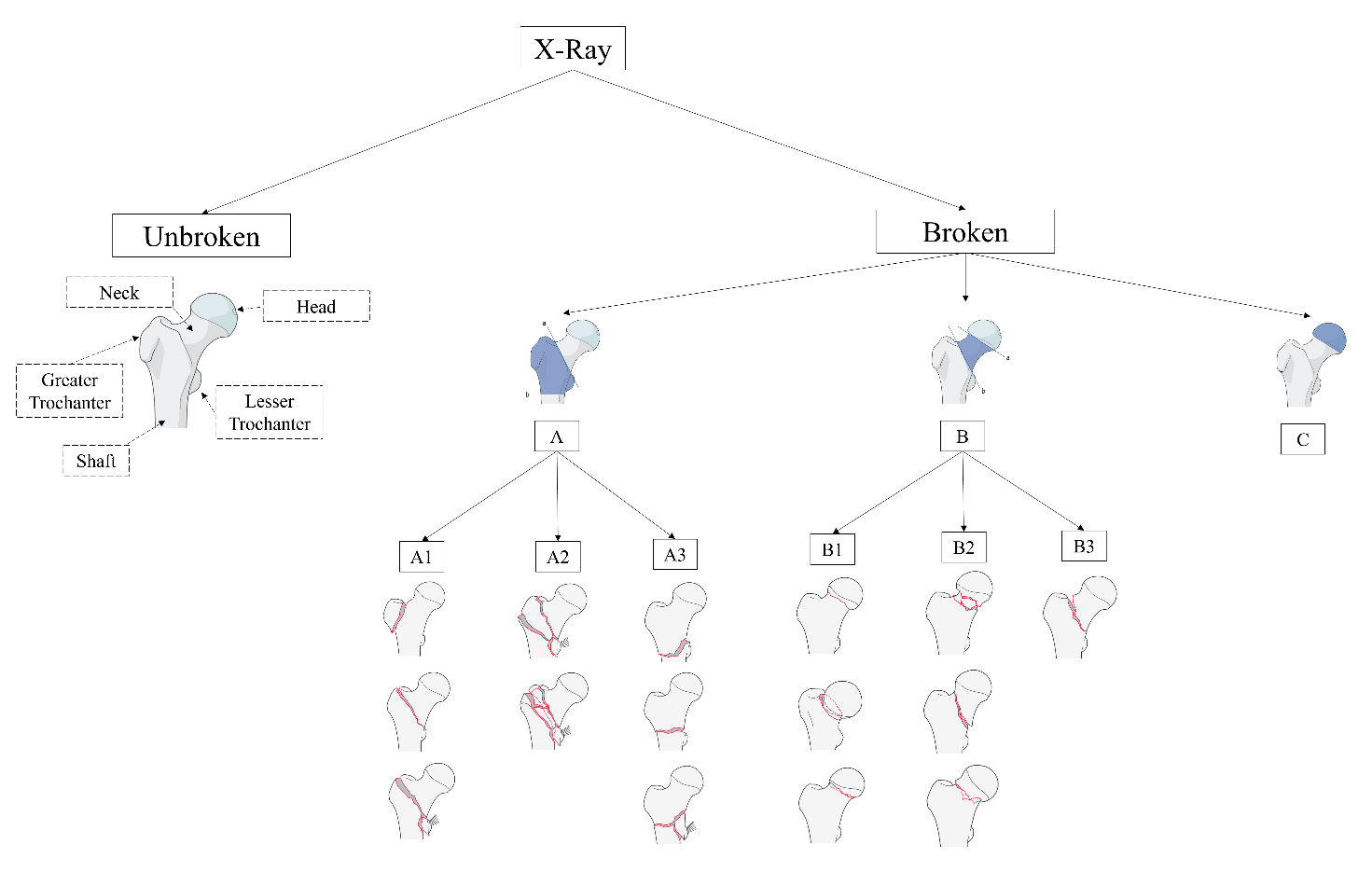}
	\caption{AO/OTA hierarchical classification determined by the localization and configuration of the fracture lines. Type \textit{A}, type \textit{B} and type \textit{C} fractures concern the trochanteric region, the femoral neck and the femoral head, respectively. Each group is then subsequently divided in different levels of subgroups}
	\label{fig:fig1}
\end{figure}

The proximal femur is labelled as “31”, being the first number related to the femur and the second to the proximal region. Then the intertrochanteric region is coded as “A”, the neck of femur “B” and the femoral head as “C”. Then the fracture is further described in groups and subgroups depending on the complexity and degree of displacement. The classification process adopted in this study is shown in \textbf{Figure \ref{fig:fig1}}.

\newpage

\subsection{Patients selection and Dataset}
\begin{figure}
	\centering
	\includegraphics[width=.8\textwidth]{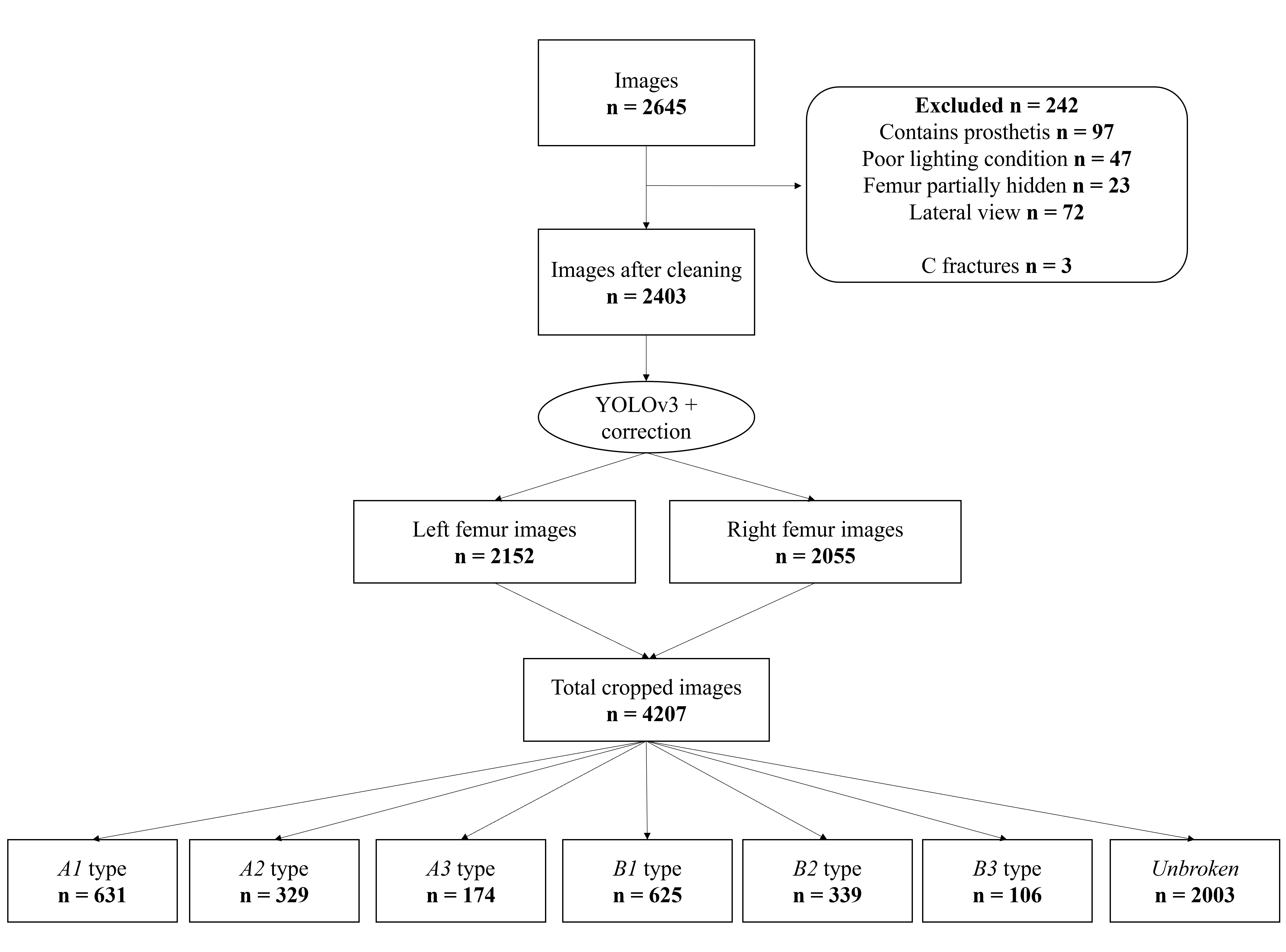}
	\caption{The STARD 2015 Flow Diagram shows the dataset processing workflow}
	\label{fig:fig2}
\end{figure}

\begin{figure} [b]
	\centering
	\includegraphics[width=.6\textwidth]{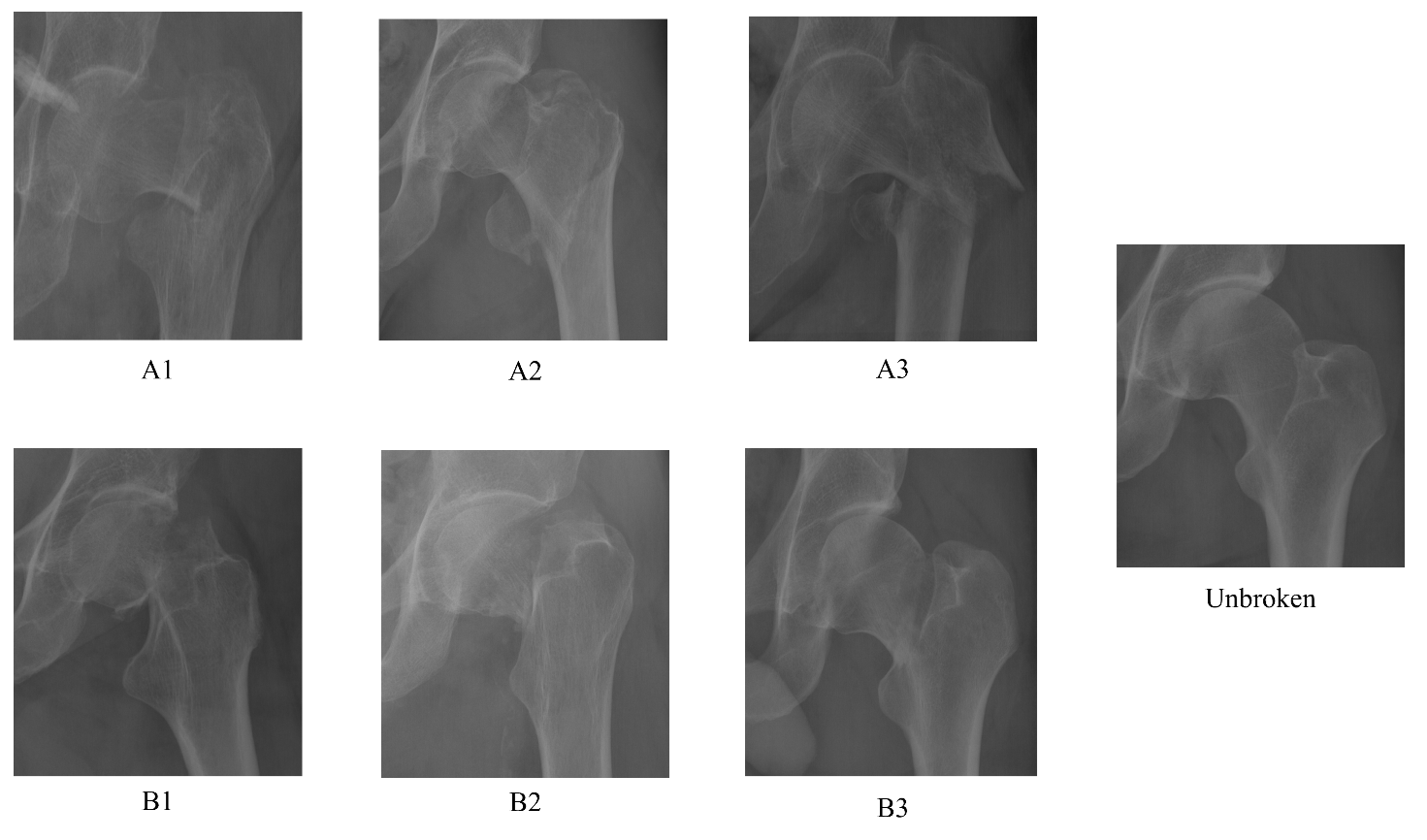}
	\caption{Some samples of real X-Rays images used for training the neural network after the cleaning and cropping phase}
	\label{fig:fig3}
\end{figure}

This retrospective study was conducted in a Level-I trauma center and was approved by . the appropriate Ethics Review Board. All >18 year-old patients with proximal femur fracture in the Emergency Department between January 2013 and December 2020 were eligible for enrollment. Exclusion criteria included missing pelvic anteroposterior radiograph documenting the hip fracture on the hospital’s PACS and pathologic fractures. Demographic data, pelvic anteroposterior radiographs with related radiological referral and intraoperative diagnosis were recorded in a computerized dataset. All data was collected anonymously using Synapse 3D (FUJIFILM Corporation) and each image was carefully examined for removal of all labels.  Table 1 describes patients’ baseline characteristics. The initial dataset was labelled by a senior trauma surgeon with 18 years of experience, a specialist who has worked specifically on femur fractures in the past 6 years and was composed by n=2645 images of the entire or half hip bone. The dataset was then reviewed using the included radiological referrals and intraoperative diagnosis. All images with discordant classification were analyzed and removed if univocal interpretation by the whole team was not attained. \textbf{Table.\ref{tab:tab1}} describes patients’ baseline characteristics.
\begin{table} [b]
	\centering
	\includegraphics[width=.6\textwidth]{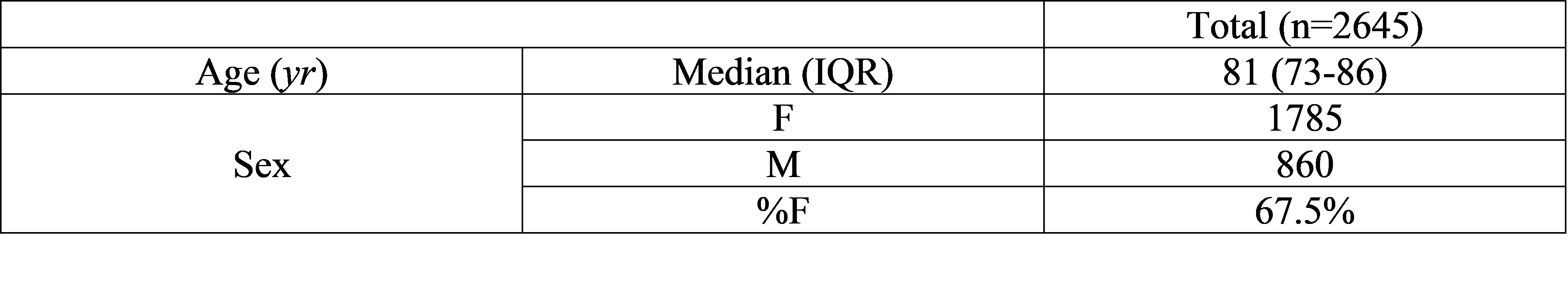}
	\caption{Baseline characteristics. Median computed with Interquartile Range (IQR)}
	\label{tab:tab1}
\end{table}
The initial dataset was labelled by a senior trauma surgeon with 18 years of experience, a specialist who has worked specifically on femur fractures in the past 6 years and was composed by n=2645 images of the entire or half hip bone. The dataset was then reviewed using the included radiological referrals and intraoperative diagnosis. All images with discordant classification were analyzed and removed if univocal interpretation by the whole team was not attained. The first step was the segmentation of the images into left and right hips. Then segmentation of the images into left and right hips was performed. The second step was a cleaning phase, where a total of \textit{n=242} images were excluded because contained prosthesis (\textit{n=97}), had poor lighting condition (\textit{n=47}), showed the area around the femur partially hidden (\textit{n=23}) or presented a lateral view (\textit{n=72}). Moreover, because of the low number of C fractures (\textit{n=3}) detectable on pelvic radiographs, this class was excluded. The third step was a cropping phase, where the areas related to the right and left femur were selected through a fully-automated cropping method and resized to $224 \times 224$, considering the fact that some images present only one between the right and left femur. This technique concerned the use of YOLOv3 (You Only Look Once) \cite{redmon_yolov3_2018} algorithm for detection of left and right femur. The fourth step was the revision of the YOLOv3 errors: \textit{n=25} femur was not detected, n=1845  right femur were correctly detected while \textit{n=208} wrongly detected as left, \textit{n=1874} left femur were correctly detected while \textit{n=241} wrongly detected as right. After the correction and the manual cropping of the not detected images, the final dataset was composed by \textit{n=2152} left femur and \textit{n=2055} right femur, which have been flipped horizontally. The fractures were then divided in different types and, afterwards, the dataset was reviewed by two radiologists from our medical team, to confirm the validity of the ground truth. The final number of images was \textit{n=4207} manually annotated images divided in different fracture types: 2003 \textit{Unbroken} femur, 631 type \textit{A1}, 329 type \textit{A2}, 174 type \textit{A3}, 625 type \textit{B1}, 339 type \textit{B2}, 106 type \textit{B3}. This process is shown in \textbf{Figure \ref{fig:fig2}} following the STARD 2015 flow diagram \cite{cohen_stard_2016}. Some real X-Rays for each class taken from our dataset are shown in \textbf{Figure \ref{fig:fig3}}. 

\subsection{ViT Configuration}
In this paper, the ViT proposed in \cite{dosovitskiy_image_2021} was applied. This architecture focuses on small patches of the image. Each patch in the input image is flattened using a linear projection matrix, and a positional embedding is added to it. The Transformer encoder, similarly to the original 2017 version [13], consists of multiple blocks of self-attention, normalization and fully connected layers with residual connections. In each attention block, multiple heads can capture different connectivity patterns. The fully connected Multi-Layer Perceptron (MLP, \cite{aggarwal_neural_2018}) head at the classification output provides the desired class prediction. As stated in the original ViT paper, this network typically requires a larger dataset than usual, as well as a longer pre-training schedule. For this reason, with only 4207 images would be unfeasible to train ViT from scratch. As a first solution, a Compact Convolutional Transformer \cite{cct} was applied, a very recent architecture, based on ViT, where the patch extraction phase is substituted by a CNN which took care of the features extraction. This solution is proven to usually overcome the big data problem, but, unfortunately, it did not applied to this case. For this reason, the four pre-trained ViT architectures were tested. In \cite{dosovitskiy_image_2021}, the proposed configurations depend on several network parameters (such as the number of neurons of a specific layer) and patches number. After experimenting with the so-called \textit{base-16}, \textit{base-32}, \textit{large-16} and \textit{large-32} configurations, the large-16 ViT block was selected, which has a multilayer perceptron of 4096 units, 16 heads, 24 layers, a hidden size of 1024 and operates with $16 \times 16$ patches. The comparison between these four configurations and the CCT are shown for completeness in \textbf{Table \ref{tab:tabnew}}.  A dense layer with 4096 neurons was added to this block, with a GELU (Gaussian Error Linear Units) activation followed by a batch normalization layer and a dropout layer with 0.5 as \textit{keep-in} parameter. Finally, a Softmax layer for 7-class classification was attached. The learning rate was initially set to $1e^{-4}$ and reduced by a factor of 0.2 after 4 epochs of plateau until a minimum of $1e^{-6}$. The optimizer used was the Rectified Adam and the loss function the categorical crossentropy. To cushion the problem of class imbalance, three diverse methods have been tried: the first assigned different weights to each class during training, where the weights are inversely proportional to the number of samples in the respective class; the other ones applied oversampling or data augmentation, with a rotation range of 10 degrees, and both height and width shift from 0.0 to 0.1 fraction of total height or width, in order to obtain the same number of samples for each class. After testing, oversampling resulted as the most performing choice. The problem with data augmentation is the fact that more complex transformations as shearing can not be used, as it may lead to the generation of “fake” fractures. To compare the results with the state of the art, two other approaches were used as baselines. The first is an InceptionV3 network, which has been chosen for the results obtained in \cite{tanzi_hierarchical_2020}. The second is a hierarchical approach, proposed in \cite{tanzi_hierarchical_2020}, which consisted in a cascade of three stages: the first network recognized \textit{Unbroken} or \textit{Broken}, the second one classified the images predicted as \textit{Broken} by the first network as \textit{A} and \textit{B} and the third and fourth ones took care of the \textit{A} and \textit{B} subgroups.

\begin{table}
	\centering
	\includegraphics[width=.8\textwidth]{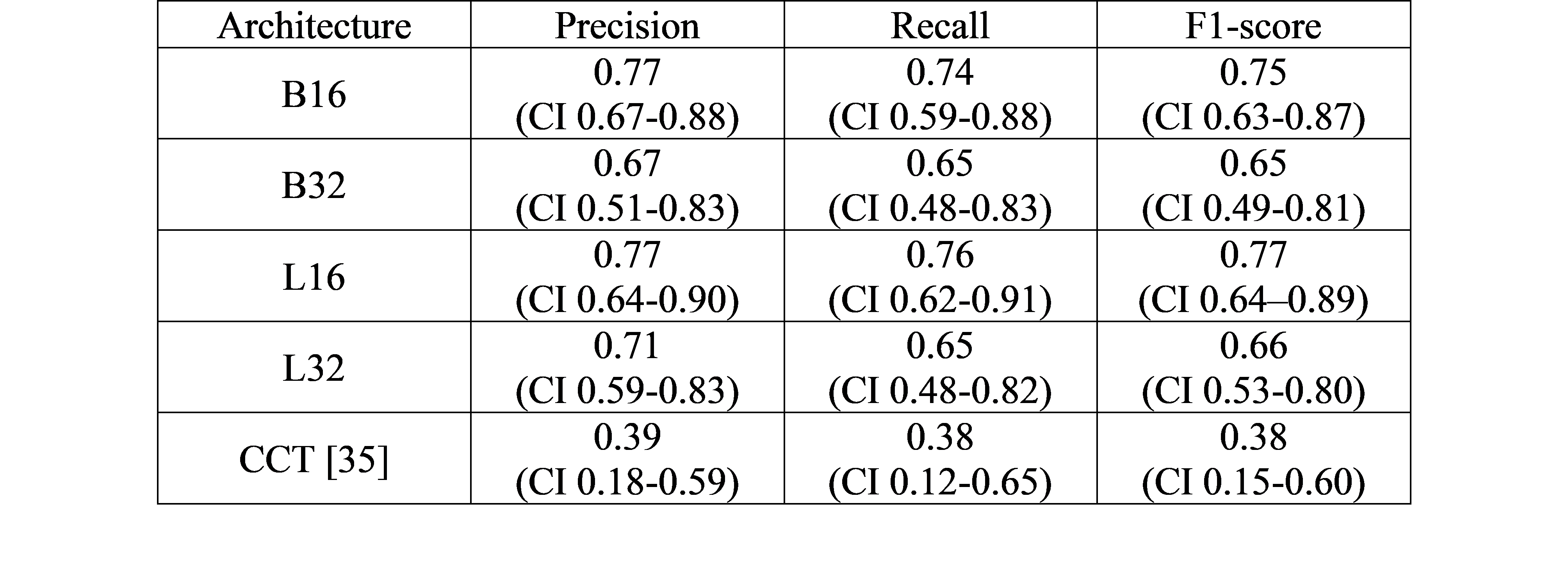}
	\caption{Values of precision, recall and F1-score for the Compact Convolution Transformer and the \textit{base-16}, \textit{base-32}, \textit{large-16}, \textit{large-32} configurations. The values are shown with related confidence interval.}
	\label{tab:tabnew}
\end{table}

\subsection{Visualization and Clustering}
For understanding where the ViT network was focusing during inference, a function was implemented in order to visualize the attention map. Then, unsupervised learning was used to evaluate the ability of the ViT encoder to extract features. Three clustering approaches were tested: firstly, the initial dataset of images resized to $224 \times 224$ was clustered using a Convolutional Autoencoder and the results were used as a baseline. Secondly, the Convolutional Autoencoder was substituted by an Autoencoder which took as input a vector of 1024 values, extracted in one case from the InceptionV3 network and in the other from the ViT encoder. In all three cases, the Autoencoder was pre-trained for 200 epochs. After this, the encoders were extracted from the three architectures and a clustering layer was added at the end. This layer was initialized with the centers found by the \textit{kmeans++} function and trained until convergence. It output a vector that represents the probability, calculated with Student's t-distribution, of the sample belonging to each cluster. The clustering performance was measured with accuracy, Normalized Mutual Information (NMI) between the ground truth and the predicted distribution, where 0 means no mutual information and 1 means a perfect correlation, Adjusted Rand Index (ARI), which computes a similarity measure by considering all pairs of samples and counting pairs that are assigned in the same or different clusters in the predicted and true clustering, and loss.

\subsection{Training, Framework and Evaluation}
From the initial dataset, 15\% of images for each class where kept apart for testing, resulting in a test set of 91 images of type \textit{A1}, 94 type \textit{A2}, 25 type \textit{A3},  90 type \textit{B1}, 49 type \textit{B2}, 16 type \textit{B3}, and 282 \textit{Unbroken} femur, and 15\% for validation. The remaining 70\% images were used for training. The networks were then trained for 40 epochs using \textit{Early Stopping} with a patience of 10 epochs. We used Keras \cite{chollet_keras_2015}, an open-source neural-network library written in Python, running on top of TensorFlow \cite{martin_abadi_tensorflow_2015}, on Windows 10 Pro with NVIDIA Quadro RTX 6000. For each network, the macro and weighted accuracy was computed. Then, the performance for single classes was measured using precision, recall and F1-score. Performance of the specialists with and without the system was computed using an online survey (Forms, Microsoft Corporation, Redmond USA) and was measured using accuracy. In this case the accuracy was a reliable metric as the dataset used was balanced. Firstly, 11 specialists (7 residents and 4 radiologists) evaluated 150 hips without the help of the neural network. The CAD tool was designed to suggest the fracture classification for the proposed image, the confidence level in percentage points and the attention map. This set of images was taken from the test dataset, and therefore not involved in the training process for obtaining comparable results. Fourteen days later, in order to produce unbiased results, the evaluation was made again but this time the specialist could consult the prediction of the ViT and the probability that the network assigned to each class.

\section{Results}

\subsection{Baseline Method}
The aforementioned InceptionV3 model was able to correctly classify 67\% of the images and obtained a macro average accuracy of 0.52 (CI 0.33-0.72), a precision of 0.57 (CI 0.42-0.72), a recall of 0.53 (CI 0.33-0.72) and a F1-score of 0.54 (CI 0.36-0.71). The hierarchical network composed by different InceptionV3 network in cascade was able to correctly classify 61\% of the images and obtained a macro average accuracy of 0.41 (CI 0.13-0.68), a precision of 0.44 (CI 0.21-0.68) , a recall of 0.41 (CI 0.14-0.69) and a F1-score of 0.40 (CI 0.15–0.64). These values are shown in \textbf{Table.\ref{tab:tab2} (a)} and \textbf{(b)}.
\subsection{ViT}
Our configuration of the ViT was able to correctly classify 83\% of the entire test dataset. The macro average accuracy obtained was 0.77 (CI 0.62-0.9), while the value of precision, recall and F1-score were 0.77 (CI 0.64-0.90), 0.76 (CI 0.62-0.91) and 0.77 (CI 0.64–0.89) respectively. These values are shown in \textbf{Table.\ref{tab:tab2} (c)}.

\begin{table}
	\centering
	\includegraphics[width=.8\textwidth]{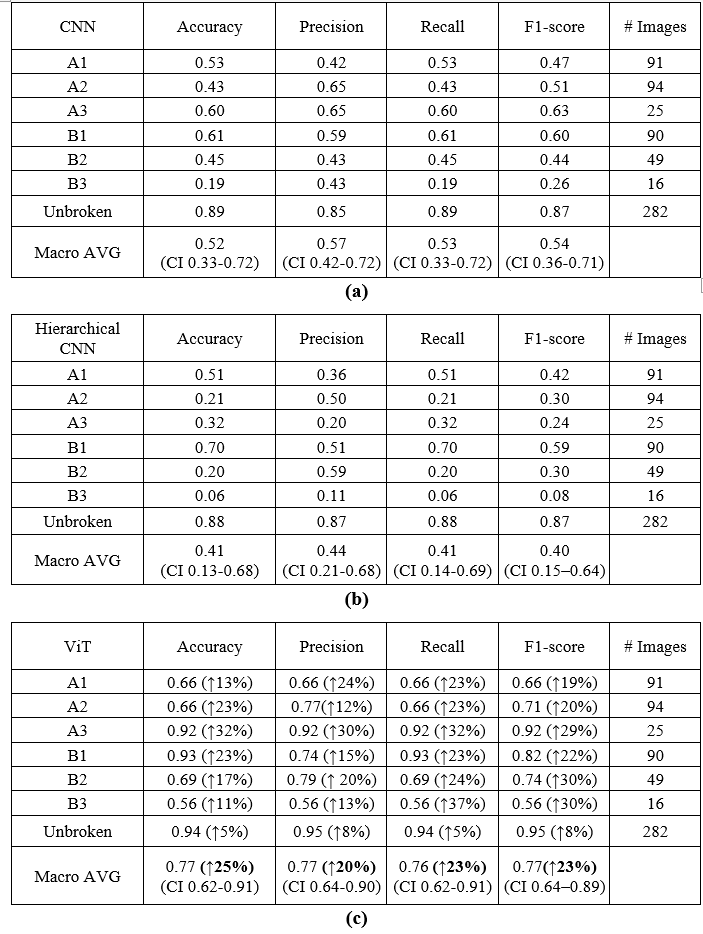}
	\caption{Values of accuracy, precision, recall and F1-score for the two baselines (a) and (b) and for the ViT (c). The values are shown with related confidence interval. In (c), the improvement given by the ViT compared to the highest value among the two baselines is shown in parenthesis.}
	\label{tab:tab2}
\end{table}

\newpage

\subsection{Clustering}
\begin{figure}[h]
	\centering
	\includegraphics[width=.6\textwidth]{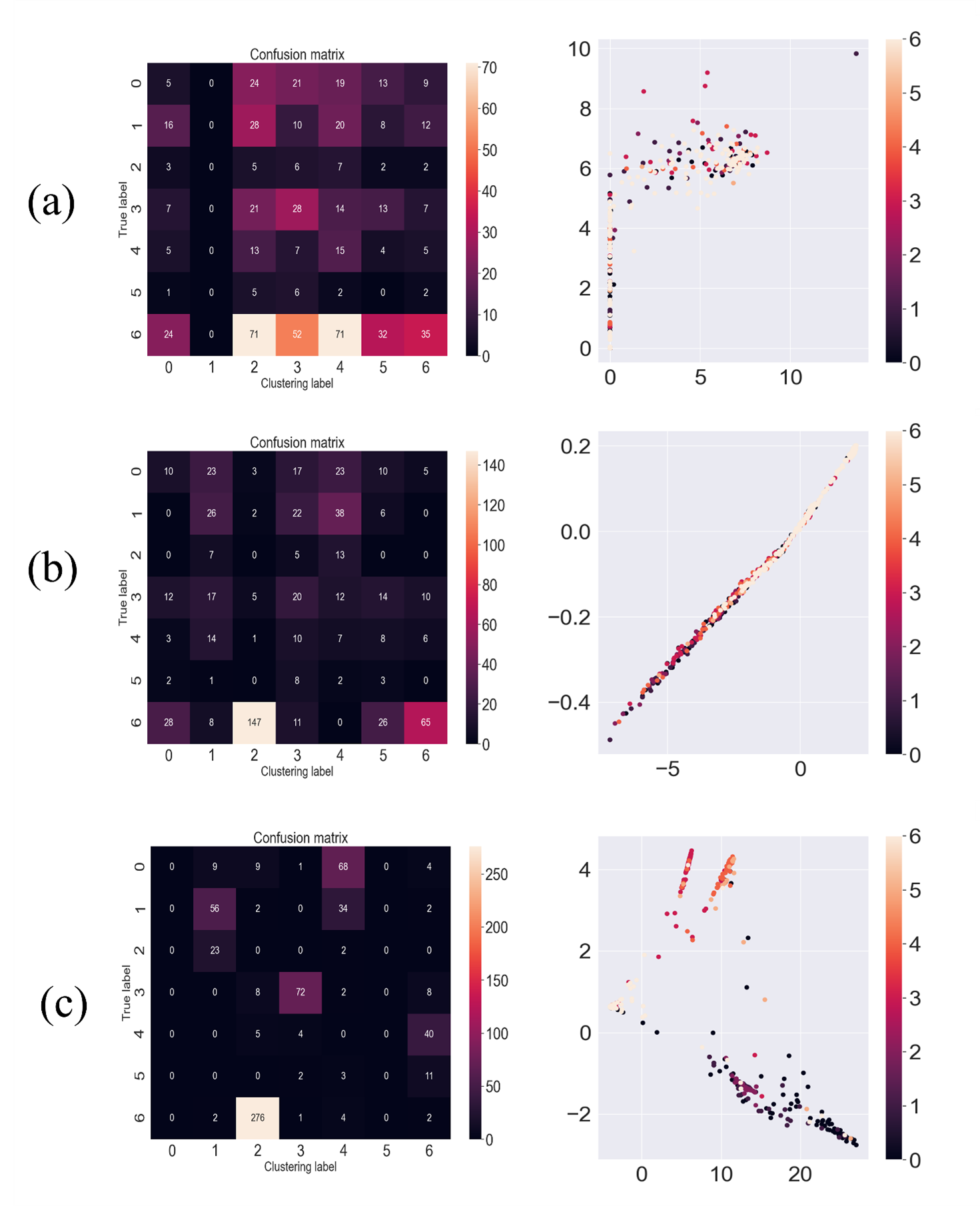}
	\caption{Confusion matrices and distributions of the clustering given by a Convolutional Autoencoder applied directly on the images (a) or by an Autoencoder applied to the feature vector extracted from the CNN (b) and extracted from the ViT Encoder (c). In the right images the cluster labels are shown with the colors presented in the sidebar}
	\label{fig:fig4}
\end{figure}

The results for the three clustering in terms of accuracy, NMI, ARI and loss are shown in \textbf{Table.\ref{tab:tab3}}, while a graphical representation of confusion matrix and distribution of each approach is shown in \textbf{Figure \ref{fig:fig4}}. 

\begin{table}
	\centering
	\includegraphics[width=.6\textwidth]{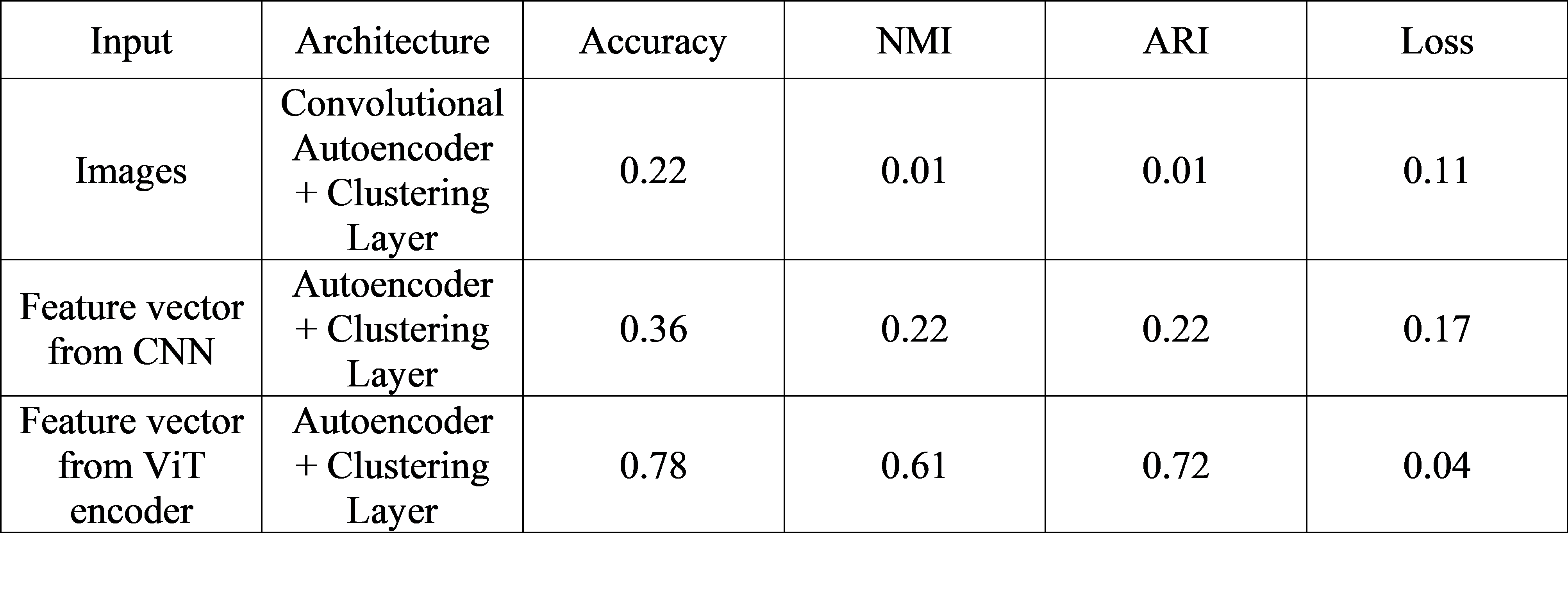}
	\caption{Values of accuracy, Normalized Mutual Information (NMI), Adjusted Rand Index (ARI) and loss for the three clustering approaches.}
	\label{tab:tab3}
\end{table}

\subsection{CAD System}
The 7-class evaluation of the type of fracture present in 150 images without the help of ViT, performed by 11 specialists and resumed in \textbf{Table.\ref{tab:tab4}}, resulted in an average accuracy of 0.58 (CI 0.53 – 0.65) for residents and 0.84 (CI 0.77 – 0.92) for radiologists. Fourteen days later, the same test was performed with the help of ViT, which with this particular set of images obtained an accuracy of 0.90 (CI 0.80-0.99). In this case, the accuracy of both residents and radiologists augmented to 0.96 (CI 0.92 – 0.99) and 1.00, respectively. The result is an average improvement of 0.29 (CI 0.12 – 0.37) in accuracy. 

\begin{table} [h]
	\centering
	\includegraphics[width=.8\textwidth]{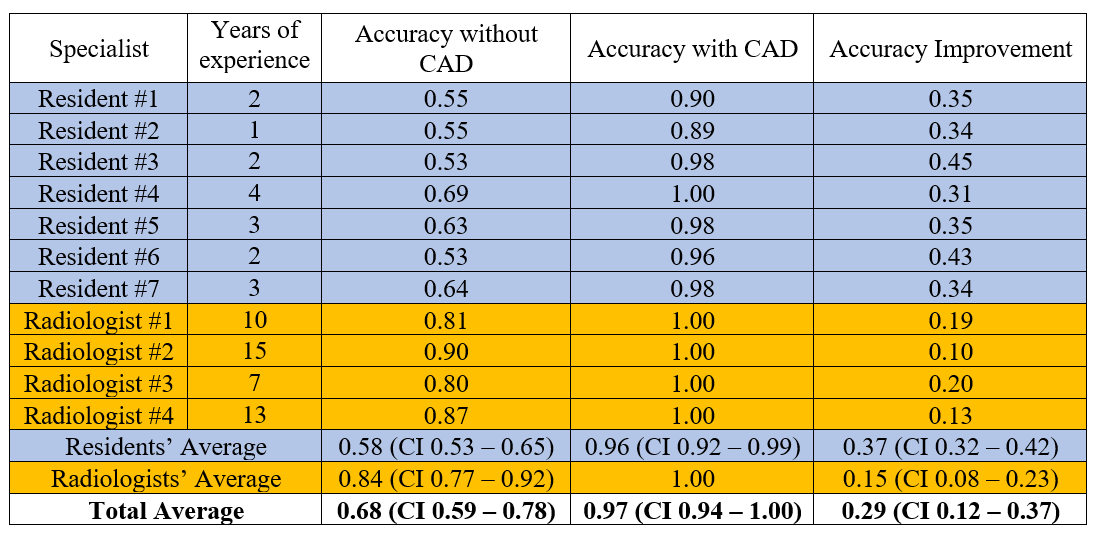}
	\caption{Values of accuracy for the 11 specialists who performed the evaluation with and without the CAD (Computer Assisted Diagnosis) system. The average values are shown with related Confidence Interval (CI)}
	\label{tab:tab4}
\end{table}

\section{Discussion and Conclusion}
\begin{figure} [h]
	\centering
	\includegraphics[width=.9 \textwidth]{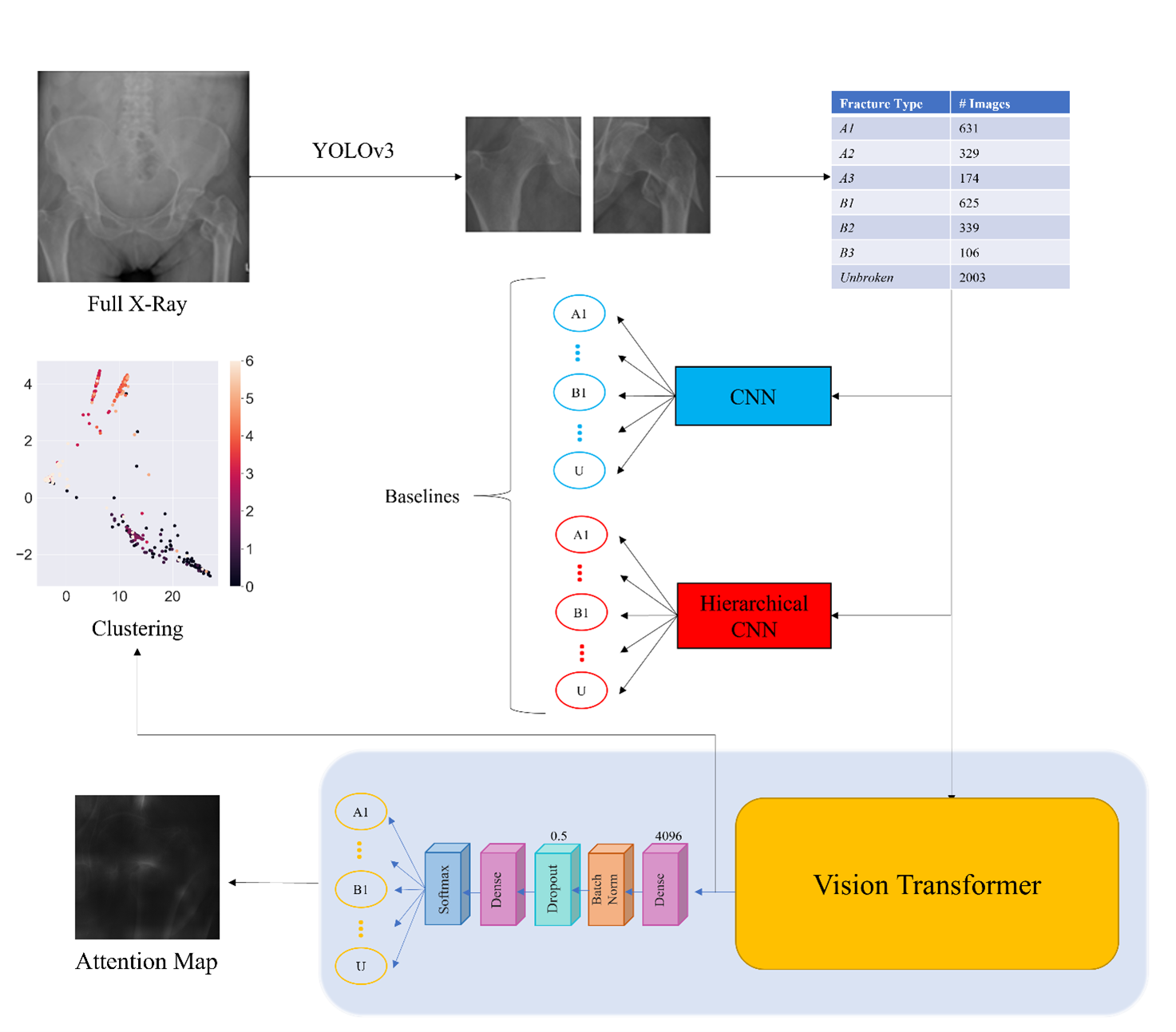}
	\caption{Full pipeline of this work. The images were firstly passed through a YOLOv3 network which cropped the areas related to the left and right femur. A CNN and a hierarchical CNN were then applied to this dataset and the results were used as baselines. Then, a modified ViT encoder was used for 7-class classification and the attention maps were analyzed. Finally, clustering was performed to evaluate the ability of the ViT to extract features}
	\label{fig:fig5}
\end{figure}

\begin{figure}
	\centering
	\includegraphics[width=.8\textwidth]{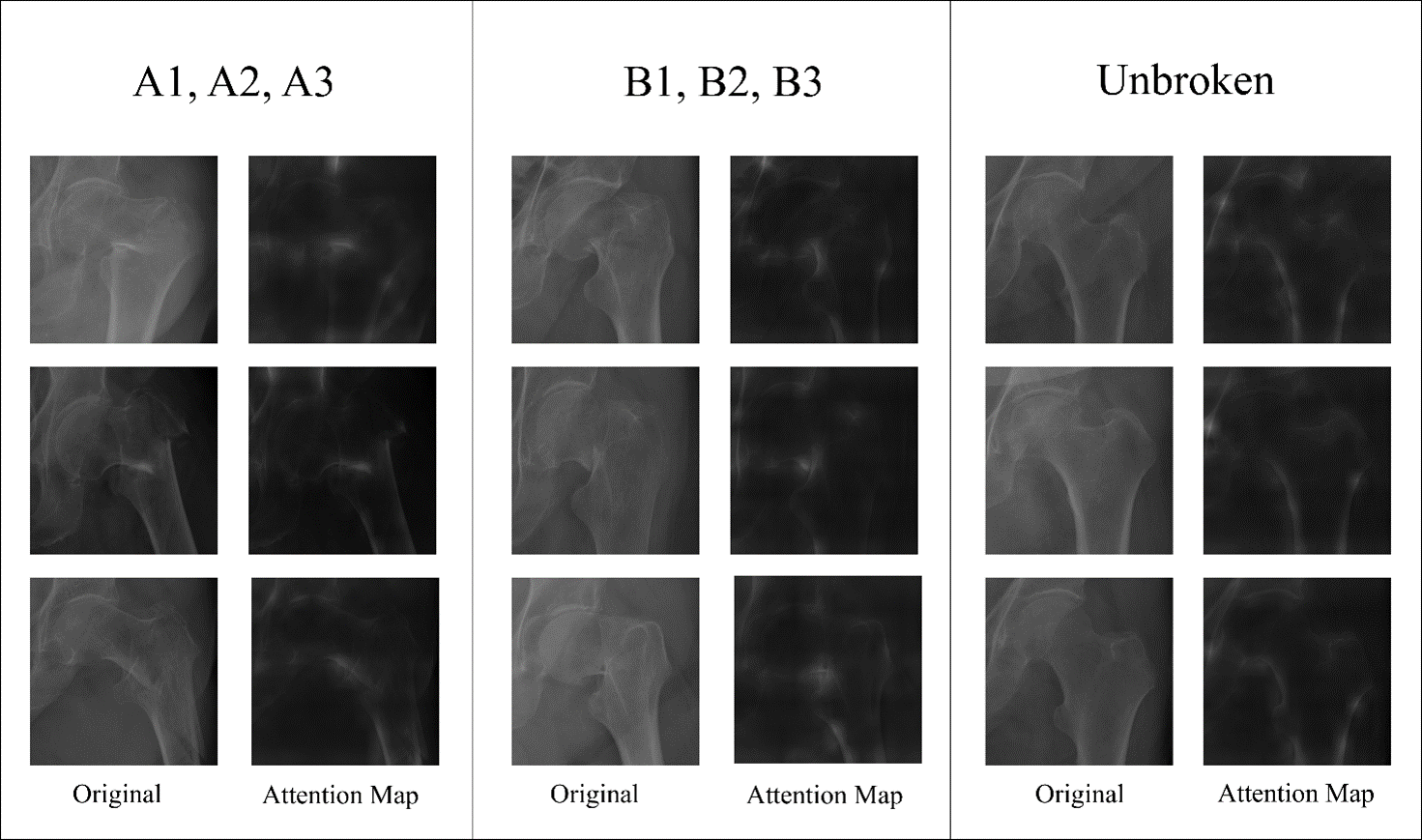}
	\caption{Some examples of original images and corresponding attention map for each class}
	\label{fig:fig6}
\end{figure}

The work introduced in this paper is summarized in \textbf{Figure \ref{fig:fig5}}. 
Firstly the original images were cropped and ordered using a YOLOv3 algorithm, obtaining a very large dataset of 4207 images divided in 7 classes. Two baselines were defined using InceptionV3 and a hierarchical network which were able to correctly classify 67\% and 61\% of the test samples, respectively. A ViT based architecture was then applied to correctly classify 83\% of the test images. The average value of accuracy, precision, recall and F1-score, resumed in \textbf{Table.\ref{tab:tab2}}, improved by a factor of 25\%, 20\%, 23\% and 23\% compared to the best result among the two baselines. The ViT was able to obtain for the first time good performance in sub-fracture classification. The attention maps in \textbf{Figure \ref{fig:fig6}} shows how the network focuses on the calcar and trochanteric area for the \textit{A} class, the neck of femur and the cortex of the greater trochanter for the \textit{B} class, and alongside the whole cortex profile of the proximal femur for the \textit{Unbroken} class. A clustering phase was also used to demonstrate how better the ViT extracts a feature vector compared to the other two approaches, even if it still struggles with some sub-fractures, as shown in \textbf{Figure \ref{fig:fig4} (c)}, where \textit{A}, \textit{B} and \textit{Unbroken} class were correctly clustered but \textit{A1}, \textit{A2}, \textit{A3} and \textit{B1}, \textit{B2}, \textit{B3} were often mismatched. Nevertheless the accuracy of this clustering was 0.78 compared to 0.22 and 0.36 of the other two (\textbf{Table.\ref{tab:tab3}}). Finally, the evaluation of 7 residents and 4 radiologists, with or without the ViT, improved by a factor of 37\% (accuracy: 0.96) and 15\% (accuracy: 1.00), respectively. These values seem excessive, but it has to be considered the fact that the specialist had access also to the probabilities, returned by ViT, related to each class. This information allowed them to focus much more on the images with uncertain probabilities and to consider also the second class predicted by ViT, in order of probability. On top of that, the best results were achieved through the synergic effect between physicians and the CAD system, resulting even better than the physicians or the algorithm alone. As already stated in \cite{doi_computer-aided_2007}, the performance of algorithms should be complementary to that by doctors rather than comparable or better with respect to them. 
The majority of the previous work concerning femur fracture classification focused on the classification between broken and unbroken bones. The clinical significance of a CAD system able to classify fractures is related to sub-fractures classification, often mismatched even by specialist with many year of experience. The two papers which defines the state of the art in femur sub-fracture classification are \cite{lee_classification_2020} and \cite{kazi_automatic_2017}. In the first, the authors collected 786 anterior-posterior X-ray images together with 459 radiology reports. The dataset was unbalanced, resulting in a very scarce test dataset. For example, just 1, 6, and 8 samples, respectively, were used to validate classes \textit{B3}, \textit{B1} and \textit{A3}. During training, the images were passed through an InceptionV3 encoder which extracted a latent representation. This latent representation was then passed to a Fully Connected (FC) classification layer and to a LSTM based decoder together with textual data from the reports. With this configurations, the authors obtained an average F1-score of 0.50, 0.27 lower than the one obtained by ViT. In addition, the approach proposed in this work does not leverage on text annotations, which are usually very hard to collect. In the second paper, the authors used an attention module to automatically locate the proximal femur area, followed by an InceptionV3 network for classification. In this case the dataset used was larger, with a total of 1173 X-Ray images, but still unbalanced, with, as the authors also underlined, as little as 15 cases for \textit{A3 fractures}. For 6 classes classification (as the class Unbroken was not considered) they obtained an average value of 0.68 for all three metrics used (precision, recall and F1-score), compared to the values of 0.77, 0.76 and 0.77, respectively, of ViT.
In summary, the results achieved with ViT outperformed the state of the art with a dataset eight times and four times larger, respectively. This aspect may seem a negative side, since it is always better to achieve certain results with as few images as possible. However, given the variety and diversity of fractures’ patterns, a larger dataset could bring to a better generalization.
On the other hand, two main problems have still to be tackled. Firstly, the unbalancing of this dataset was bypassed with oversampling, but a better approach would be to augment the dataset using Generative Adversarial Networks \cite{goodfellow_generative_2014} to produce new and reliable samples \cite{frid-adar_synthetic_2018}. The problem with GANs is that they might not work well with classes which present a low number of samples, as, fewer the images available for training, less their ability generalize and create new images will be. For this reason, a solution could be to produce new generic samples, training the architecture with the entire dataset, and ask the specialists to classify the generated images. This idea will be discussed in a future paper. Secondly, even if in this paper the hierarchical approach yielded very low results, the performance of the ViT could massively improve adapting its architecture to the hierarchical structure of the AO/OTA classification.
In conclusion, this is the first work where a ViT architecture was applied to femur fracture classification. It  outperformed the state of the art approaches based on CNN. In future work, this method will be improved and extended for even more complex levels of sub fractures in the AO/OTA classification.

\newpage

\UseRawInputEncoding
\bibliographystyle{unsrtnat}
\bibliography{references}

\end{document}